# DemandLens: Enhancing Forecast Accuracy Through Product-Specific Hyperparameter Optimization

Srijesh Pillai
Department of Computer Science & Engineering
Manipal Academy of Higher Education
Dubai, UAE
srijesh.nellaiappan@dxb.manipal.edu

M. I. Jawid Nazir
Department of Computer Science & Engineering
Manipal Academy of Higher Education
Dubai, UAE
jawid_nazir@manipaldubai.com

*Abstract* — DemandLens demonstrates an innovative Prophet-based forecasting model for the mattress-in-a-box industry, incorporating COVID-19 metrics and SKU-specific hyperparameter optimization. This industry has seen significant growth of E-commerce players in the recent years, wherein the business model majorly relies on outsourcing Mattress manufacturing and related logistics and supply chain operations, focusing on marketing the product and driving conversions through Direct-to-Consumer sales channels. Now, within the United States, there are a limited number of Mattress contract manufacturers available, and hence, it is important that they manage their raw materials, supply chain, and, inventory intelligently, to be able to cater maximum Mattress brands. Our approach addresses the critical need for accurate Sales Forecasting in an industry that is heavily dependent on third-party Contract Manufacturing. This, in turn, helps the contract manufacturers to be prepared, hence, avoiding bottleneck scenarios, and aiding them to source raw materials at optimal rates. The model demonstrates strong predictive capabilities through SKU-specific Hyperparameter optimization, offering the Contract Manufacturers and Mattress brands a reliable tool to streamline supply chain operations.

## I. Introduction

The Mattress-in-a-Box industry in the United States has experienced substantial growth, especially during the COVID-19 pandemic. Currently valued at approximately $20 billion, with a Compound Annual Growth Rate (CAGR) of around 4%, the business model involves Direct-To-Consumer sales through E-commerce platforms (Websites, etc.), with compressed mattresses shipped in compact boxes. This approach eliminates the need for large warehouses and reduces logistical expenses significantly [1].

A distinctive feature of this industry is its heavy reliance on third-party Contract Manufacturers. Most mattress brands outsource production to specialized Mattress manufacturing facilities, focusing their internal resources on Marketing Analytics, Customer Acquisition, and Conversion Rate Optimization. This model creates a critical dependency: manufacturers require accurate sales forecasts to plan production capacity, secure raw materials, and optimize workforce allocation.

With limited world-class manufacturing facilities serving numerous brands within the Mattress industry in the United States, accurate demand forecasting becomes essential for both parties. Brands must provide reliable monthly estimates to secure manufacturing capacity, while manufacturers need these forecasts to efficiently allocate resources across multiple clients (Mattress brands). The coordination challenge is further complicated by the variety of mattress sizes (10-inch, 12-inch, 14-inch, 16-inch) and variants (Twin, Twin XL, Full, King, Queen, Cal King, and Split King), each with distinct sales patterns and seasonal variations.

This research presents a forecasting model developed for the Mattress-in-a-Box industry, tested and validated on real sales data of a luxury Mattress brand in the US market, addressing these unique challenges through advanced time series analysis enhanced with COVID-19 impact variables.

## II. Literature Review

In August 2024, we conducted an extensive review of demand forecasting methodologies in the Mattress industry, examining both traditional and emerging approaches. Our investigation identified several factors influencing Mattress sales patterns that directly impact demand forecasting accuracy:

1. Economic Indicators: Consumer confidence index, unemployment rates, and interest rates have shown significant correlation with purchasing decisions for home-related products. Our investigation revealed that a 1% change in consumer confidence can influence demand by up to 3.2% in subsequent months [1].

2. Supply Chain Factors: Material availability, manufacturing capacity constraints, and shipping delays create significant variability in product availability. The post-pandemic era has introduced persistent supply uncertainties, with lead time fluctuations ranging from 2-8 weeks for similar products [2].

3. Online Search Trends: Web search volume (online search keywords) demonstrates strong predictive

value, typically preceding purchase decisions by 2-4 weeks. Analysis of keyword trends provides early warning signals for demand shifts, with correlation coefficients of 0.72-0.85 between search intensity and subsequent sales [3].

4. Price and Promotions: Discount depth and promotional timing create substantial demand spikes that disrupt normal patterns. Major E-commerce platforms observe demand elasticity ranging from 1.8-3.2 depending on product category and price point, requiring specialized modelling approaches that account for both demand forecasting and price optimization dynamics in an online retail context [4][5].

5. Competitor Activity: Competitor actions, particularly price changes and marketing campaigns, can rapidly shift market share. Our analysis observed market share volatility of up to 15% within a 30-day period following major competitor initiatives [6].

6. Seasonal Trends: Distinct cyclical patterns occur throughout the calendar year, with notable peaks during summer months, major holidays, and home-buying seasons. The timing and magnitude of these cycles vary by region and demographic segment [7].

Each of these factors in turn affects the demand patterns, the planning horizons, and the forecast accuracy. The combined effect of these variables creates a complex forecasting environment requiring sophisticated modelling approaches.

Our review identified a significant gap regarding the integration of COVID-19 effects into forecasting models. The pandemic caused unprecedented disruption in consumer behaviour, with home-related purchases seeing volatility 3-4 times historical norms [8]. Traditional forecasting methods struggled to incorporate these seasonal variations, highlighting the need for more adaptive approaches.

Previous studies have explored various forecasting methods for retail and manufacturing, including ARIMA models, exponential smoothing, and more recently, machine learning approaches [9][10][11]. However, there remains a gap in forecasting methodologies specifically tailored to the Mattress-in-a-Box industry with its unique direct-to-consumer channel and outsourced manufacturing model, and specifically taking into account the outliers introduced by the Pandemic era.

The Fig. 1 below shows a holistic view of the factors that must be taken into account while developing a forecasting model for the mattress industry.

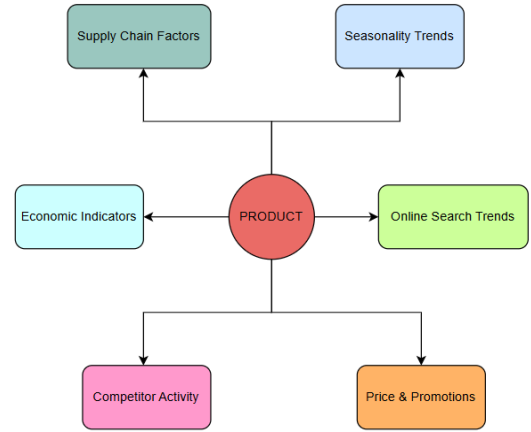

Fig 1. Factors influencing E-Commerce Demand Forecasting for the Mattress Industry

III. METHODOLOGY

The methodology for our forecasting model was developed through a systematic approach that addressed the unique challenges of direct-to-consumer product demand prediction in a pandemic-disrupted market environment. Our methodology combines traditional time series techniques with novel applications of external regressors and SKU-specific optimization.

A. Research Design

Our research followed a multi-stage approach designed to answer three central questions:

1. How can traditional time series forecasting be enhanced to incorporate pandemic-related disruptions?

2. To what extent do different product variants require customized modelling approaches?

3. What is the optimal forecast horizon for contract manufacturing coordination specifically in the Mattress industry?

We employed a mixed-methods approach combining exploratory data analysis, hypothesis testing, model development, and empirical validation using real data from a US-based luxury direct-to-consumer Mattress brand.

B. Data Selection Framework

We established rigorous criteria for the selection and preparation of two critical datasets:

1. Sales Data Requirements:
   - Historical data spanning April 2018 through July 2024
   - Day-wise orders granularity
   - Complete SKU differentiation (10-inch, 12-inch, 14-inch, and 16-inch variants)
   - Clean delineation of promotional periods

2. COVID-19 Data Requirements:

- Temporal coverage from January 21, 2020, to March 23, 2023
- Daily case counts and mortality figures
- Consistent reporting methodology

Our hypothesis linking COVID-19 mortality statistics to purchasing behaviour was based on the causal chain: increased deaths → stricter lockdowns → people spend more time at home → increased focus on home environments → higher probability of home product purchases.

C. Decomposition and Feature Engineering Approach

We developed a structured approach to feature engineering that decomposed the forecasting problem into multiple components:

1. Temporal Pattern Identification: Analysis of cyclical patterns at daily, weekly, and seasonal levels.
2. Lag Feature Design: Creation of structured time-shifted features at 1, 7, 14, 30, and 60-day intervals.
3. Rolling Window Selection: Implementation of 7, 14, and 30-day rolling averages to capture trend momentum.
4. Seasonal Flag Development: Binary indicators for known high-demand periods, such as promotions, national holidays, etc.
5. COVID-19 Signal Processing: Transformation of raw pandemic metrics into predictive indicators.

The mathematical foundation for our approach incorporated both additive and multiplicative components, consistent with established time series decomposition principles as outlined by Hyndman and Athanasopoulos [13]:

$$y(t) = g(t) + s(t) + h(t) + \sum_{i=1}^{n} \beta_i x_i(t) + \varepsilon_t \quad \ldots (1)$$

Where:
- $y(t)$ is the forecast value at time $t$
- $g(t)$ represents the trend component
- $s(t)$ captures seasonal variations
- $h(t)$ accounts for holiday effects
- $\beta_i$ are the coefficients for additional features
- $x_i(t)$ are the values of external regressors at time $t$ (including COVID-19 metrics)
- $\varepsilon_t$ is the error term

This decomposition allows for explicit modelling of each component while incorporating the impact of pandemic variables through the external regressor terms.

D. Algorithm Selection Methodology

After evaluating multiple candidate algorithms including ARIMA, exponential smoothing, and various machine learning approaches, we selected Facebook's Prophet algorithm as our foundation based on the following criteria:

1. Multiple Seasonality Handling: Ability to model overlapping seasonal patterns.
2. Changepoint Detection: Automatic identification of trend shifts.
3. Incorporation of External Regressors: Support for additional predictive variables.
4. Interpretable Components: Clear separation of trend, seasonal, and holiday effects.
5. Robust Missing Data Handling: Ability to function with incomplete historical data.

This selection was validated through comparative benchmarking against traditional statistical and machine learning alternatives, with Prophet demonstrating superior performance for our specific use case [12].

Table I. COMPARATIVE ANALYSIS OF THE CAPABILITIES OF MULTIPLE ALGORITHMS [12][13][14][15]

| Capability | ARIMA | LSTM | XGBoost | Prophet |
|---|---|---|---|---|
| Multiple Seasonality | Single seasonal component | Learns patterns implicitly | Through feature-engineering | Multiple explicit seasonalities |
| External Regressors | ARIMAX implementation available | Multi-input architecture | Native feature support | add_regressor() method |
| Interpretability | Coefficient interpretation | Limited explainability | Feature importance scores | Component decomposition |
| Trend Changes | Manual structural break detection | Adaptive learning | Captures through splits | Automatic changepoint detection |
| Missing Data | Requires complete series | Sequence handling needed | Handles missing feature | Built-in Interpolation |
| Business Application | Statistical expertise required | Deep learning expertise | ML Expertise required | Designed for Analysts |

E. Hyperparameter Optimization Strategy

We developed a methodical approach to hyperparameter tuning that recognized the fundamental differences between product variants:

1. Variant-Specific Analysis: Assessment of historical patterns for each SKU to identify unique characteristics.
2. Parameter Space Definition: Establishment of search ranges for critical parameters:
   - *changepoint_prior_scale*: Controls trend flexibility (0.001 to 0.5)
   - *seasonality_prior_scale*: Controls seasonal strength (1.0 to 50.0)
   - *holidays_prior_scale*: Controls holiday impact (1.0 to 25.0)
   - *seasonality_mode*: Additive vs. multiplicative seasonality
   - *changepoint_range*: Proportion of history where trend changes can occur (0.8 to 0.97)
   - *n_changepoints*: Number of potential trend change points (15 to 55)
3. Cross-Validation Framework: Implementation of time-based cross-validation with expanding windows.

4. Optimization Metric: Mean Absolute Percentage Error (MAPE) minimization.

This approach yielded distinct hyperparameter sets for each product variant, reflecting their unique demand dynamics and sensitivity to external factors.

F. COVID-19 Integration Method

Our methodology for incorporating pandemic effects was based on the dual use of case counts and mortality statistics:

1. Signal Selection: Analysis of correlation patterns between various pandemic metrics and demand changes.
2. Temporal Alignment: Synchronization of pandemic data with sales transactions.
3. Rolling Window Optimization: Determination of optimal smoothing intervals for raw metrics.
4. Coefficient Analysis: Evaluation of the relative importance of case vs. mortality data.

The methodology was guided by the hypothesis that mortality figures would serve as a more reliable proxy for lockdown severity than case counts alone, particularly given the variability in testing protocols throughout the pandemic.

G. Forecast Evaluation Framework

We established a comprehensive evaluation methodology to assess forecast quality:

1. Time Horizons: Primary focus on 1-month, 2-month, and 3-month forecasts
2. Error Metrics: MAPE (Mean Absolute Percentage Error), RMSE (Root Mean Square Error), and MAE (Mean Absolute Error)
3. Directional Accuracy: Assessment of correct trend prediction independent of absolute values
4. SKU-Specific Performance: Disaggregated evaluation by product variant
5. COVID-19 Sensitivity: Analysis of forecast performance during high-volatility pandemic periods

This multi-dimensional assessment provided a robust measure of model performance across different conditions and forecast horizons.

H. Manufacturing Coordination Method

Our methodology established clear linkages between forecasting outputs and manufacturing planning requirements:

1. Horizon Alignment: Design of forecast periods to match manufacturing planning cycles.
2. Aggregation Level: Monthly totals with SKU-level differentiation.
3. Lead Time Consideration: Incorporation of production and shipping lead times.
4. Confidence Interval Application: Provision of uncertainty bounds for capacity planning.

This approach ensured that the forecasting system would generate actionable outputs directly applicable to manufacturing coordination challenges.

Through this methodological framework, we developed a forecasting approach that addresses the unique challenges of pandemic-era demand prediction while providing practical value for manufacturing planning. The method combines statistical rigor with domain-specific knowledge to create a system optimized for the direct-to-consumer product environment.

IV. DESIGN & IMPLEMENTATION

Building upon our methodological framework, this section details the technical architecture and implementation choices made in our forecasting system, with emphasis on the mathematical foundations and engineering decisions.

A. System Architecture

We designed a modular forecasting system with six interconnected components that process data through a series of transformations, as illustrated in Fig. 2. This architecture follows a directed acyclic graph (DAG) pattern, ensuring unidirectional data flow while maintaining component isolation for easier maintenance and testing.

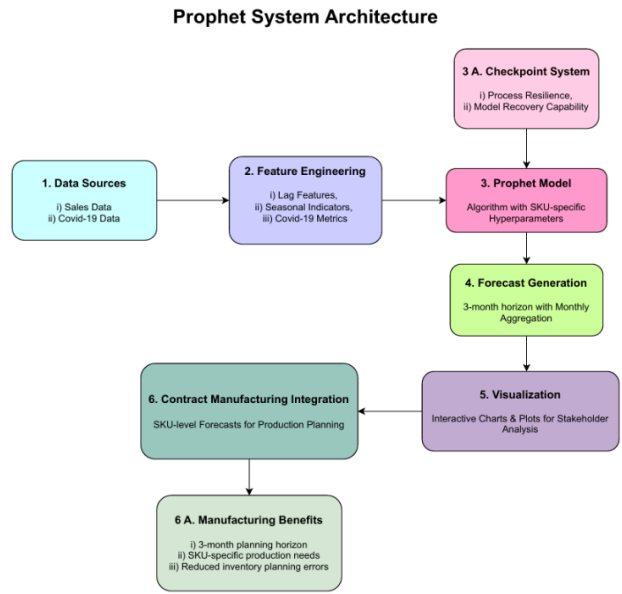

Fig 2. Our Hyper-parameter tuned Prophet Model's complete System Architecture

B. COVID-19 Data Integration

A principal innovation in our system is the integration of pandemic data as a predictive factor. The implementation focused on transforming raw COVID-19 statistics into actionable features.

```
# Add COVID features
df = df.merge(covid_data, on='dt', how='left')
df['new_cases'] = df['new_cases'].fillna(0)
df['new_deaths'] = df['new_deaths'].fillna(0)
df['cases_7day_avg'] = df.groupby('sku')['new_cases'].transform(lambda x: x.rolling(window=7).mean())
df['deaths_7day_avg'] = df.groupby('sku')['new_deaths'].transform(lambda x: x.rolling(window=7).mean())
```

Fig 3. Implementation of the Mathematical Formulation of Rolling 7 Day Average for new Covid Cases and Deaths

We specifically chose a 7-day rolling average to smooth daily fluctuations in reporting while preserving the underlying signal. The decision to use a rolling average rather than simple lag features was guided by the observation that COVID-19 reporting often exhibited day-of-week effects and reporting anomalies that could introduce noise. The mathematical formulation for our rolling average is:

$$C_{t,7} = \frac{1}{7}\sum_{i=0}^{6} c_{t-i} \quad \ldots (2)$$

$$D_{t,7} = \frac{1}{7}\sum_{i=0}^{6} d_{t-i} \quad \ldots (3)$$

Where:

- $C_{t,7}$ is the 7-day rolling average of COVID-19 cases at time t
- $D_{t,7}$ is the 7-day rolling average of COVID-19 deaths at time t
- $c_{t-i}$ represents the raw case count *i* days before time *t*
- $d_{t-i}$ represents the raw case count *i* days before time *t*

These smoothed metrics serve as indicators of lockdown intensity, which we hypothesized would correlate with changes in home product purchasing behaviour.

C. Prophet Model Implementation

The core of our forecasting engine is built on Facebook's Prophet algorithm, which decomposes time series data into trend, seasonality, and holiday components:

$$y(t) = g(t) + s(t) + h(t) + \varepsilon_t \quad \ldots (4)$$

Where:

- $y(t)$ is the predicted demand at time *t*
- $g(t)$ represents the trend component, capturing the non-periodic changes in the time series
- $s(t)$ captures seasonal variations (daily, weekly, yearly) that repeat at regular intervals
- $h(t)$ accounts for holiday effects, allowing for irregular schedules of known events
- $\varepsilon_t$ is the error term, representing unexplained variation

We extended this model by incorporating additional regressors, modifying the equation to:

$$y(t) = g(t) + s(t) + h(t) + \sum_{i=1}^{n} \beta_i x_i(t) + \varepsilon_t \quad \ldots (5)$$

Where:

- $\beta_i$ are the regression coefficients that quantify the impact of each external feature
- $x_i(t)$ are the values of our additional features at time *t*, including:
    - Lag features (*lag1, lag7, lag14, lag30, lag60*) capturing recent sales activity
    - Rolling means (*rolling_mean_7, rolling_mean_14, rolling_mean_30*) providing smoothed momentum indicators
    - Seasonal flags (*is_weekend, is_summer_peak, is_black_friday, is_back_to_school, is_holiday_season*) identifying key high-demand periods
    - Calendar indicators (quarter) capturing broader seasonal patterns
    - COVID-19 metrics (*cases_7day_avg, deaths_7day_avg*) reflecting pandemic intensity

The implementation of this extended model required careful consideration of Prophet's API [12].

```
model = Prophet(
    daily_seasonality=True,
    weekly_seasonality=True,
    yearly_seasonality=True,
    holidays=holiday_df,
    **best_params
)

for col in ['lag_1', 'lag_7', 'lag_14', 'lag_30', 'lag_60', 'rolling_mean_7', 'rolling_mean_14',
    'rolling_mean_30', 'is_weekend', 'is_summer_peak', 'is_black_friday', 'is_back_to_school',
    'is_holiday_season', 'quarter', 'cases_7day_avg', 'deaths_7day_avg']:
    model.add_regressor(col)

model.fit(product_data)
print(f"Model fitted for SKU {row['sku']}")
```

Fig 4. Implementation of feeding all our variables to the Prophet Model

Each regressor was added individually rather than in batch to ensure proper handling and to facilitate future extensions of the feature set. This approach also allows for explicit control over feature normalization, which we disabled to preserve the interpretability of coefficients.

D. SKU-Specific Hyperparameter Optimization

One of the most technically nuanced aspects of our implementation was the development of customized hyperparameter sets for each product variant. Prophet's flexibility is governed by several key parameters that control trend flexibility, seasonality strength, and changepoint detection.

```
SKU_HYPERPARAMETERS = {
    '10-inch mattresses': {
        'changepoint_prior_scale': 0.2,
        'seasonality_prior_scale': 50.0,
        'holidays_prior_scale': 25.0,
        'seasonality_mode': 'multiplicative',
        'changepoint_range': 0.97,
        'n_changepoints': 55
    },
    '12-inch mattresses': {
        'changepoint_prior_scale': 0.12,
        'seasonality_prior_scale': 40.0,
        'holidays_prior_scale': 25.0,
        'seasonality_mode': 'multiplicative',
        'changepoint_range': 0.92,
        'n_changepoints': 48
    },
```

Fig 5. Customized Hyperparameter sets for 10-inch and 12-inch variants

The changepoint_prior_scale parameter deserves particular attention as it governs the flexibility of the trend component through a Laplace prior on the rate of change:

$$\Delta g_i \sim Laplace(0, \tau)$$

Where:

- $\Delta g_i$ is the change in trend at the *i*-th changepoint
- $\tau$ is the scale parameter controlled by changepoint_prior_scale

- *Laplace(0,τ)* denotes a Laplace distribution with location 0 and scale $\tau$

Higher values of $\tau$ allow for more abrupt changes in the trend, which was necessary for the 16-inch variant that exhibited more volatile demand patterns. Conversely, the 12-inch variant, with more stable demand, benefited from a much lower value (0.01).

The decision to use multiplicative seasonality (seasonality_mode: 'multiplicative') was driven by the observation that seasonal variations scaled with the overall trend level. This is mathematically represented as:

$$y(t) = g(t) \cdot (1 + s(t)) + h(t) + \varepsilon_t \ldots (6)$$

This formulation allows seasonal amplitudes to increase as the trend increases, which aligns with the observed sales patterns in our raw day-wise data. Specifically:

- When $g(t)$ (trend) is large, the impact of $s(t)$ (seasonality) is proportionally larger
- When $g(t)$ is small, seasonal variations have less absolute impact

This corresponds to our observation that higher-selling periods exhibited larger absolute seasonal fluctuations, while maintaining similar percentage variations.

### E. Forecast Generation and Projection

A critical technical challenge in our implementation was the projection of feature values into the forecast period. For recursive features like lags and rolling averages, we implemented a forward-filling approach:

We specifically chose a 3-period rolling mean of the most recent values rather than simple forward-filling of the last value. This decision was motivated by our finding that recent averages provided more stable projections than single-point values, particularly important for recursive features that could otherwise propagate anomalies. The mathematical formulation for this projection is:

$$x_{t+h}^{(i)} = \frac{1}{3} \sum_{j=1}^{3} x_{t-j+1}^{(i)} \ldots (7)$$

Where:

- $x_{t+h}^{(i)}$ is the projected value of feature $i$ at horizon $h$
- $t$ is the last observed time point
- $j$ indexes the most recent observations
- The summation calculates a 3-period average of the most recent values

This approach reduces the impact of potential outliers in the most recent observations while providing a reasonable proxy for the expected feature values during the forecast period.

### F. Monthly Aggregation Technique

For manufacturing planning purposes, our system needed to aggregate daily forecasts to monthly totals. We implemented a precise month-difference calculation to align forecasts with manufacturing cycles:

```
forecast['month_diff'] = forecast.apply(lambda row:
                        (row['ds'].year - row['data_cutoff_date'].year) * 12 +
                        row['ds'].month - row['data_cutoff_date'].month, axis=1)
```

Fig 6. Implementation of the month-difference calculation

This calculation creates a relative month index that starts at 0 for the current month, 1 for next month, and so on. The implementation uses a vectorized approach that avoids explicit loops for performance reasons. The mathematical definition is:

$$m_{diff} = 12 * (y_{forecast} - y_{cutoff}) + (m_{forecast} - m_{cutoff}) \ldots (8)$$

Where:

- $m_{diff}$ is the month difference index
- $y_{forecast}$ is the year of the forecast date
- $y_{cutoff}$ is the year of the data cutoff date
- $m_{forecast}$ is the month of the forecast date
- $m_{cutoff}$ is the month of the data cutoff date

The aggregation uses a sum function to convert daily predictions to monthly totals, preserving the additive nature of

```
forecast_month_aggregate = forecast.groupby(['month','year','month_diff']).agg(
    sales=('yhat', 'sum')
).reset_index()
forecast_month_aggregate['sku'] = row['sku']
```

the forecast while reducing granularity to match manufacturing planning horizons.

Fig 7. Implementation of the aggregation logic to convert daily predictions to monthly totals

Through these technical implementations, we created a forecasting system that effectively integrates pandemic impacts, accommodates product-specific demand patterns, and provides mathematically sound projections for manufacturing planning purposes. The system's architecture balances computational efficiency with forecast accuracy, while the checkpoint mechanism ensures operational resilience.

## V. DISCUSSIONS

To further understand the forecasting model's implications, limitations, and potential improvements, we engaged in discussions with industry stakeholders and conducted an in-depth analysis of its practical applications. This section explores key insights gained from these discussions and examines the broader context of implementing pandemic-aware forecasting in direct-to-consumer manufacturing environments.

### A. Practical Implementation Considerations

Our discussions with industry stakeholders in the direct-to-consumer mattress industry revealed several important factors affecting real-world deployment:

1. Forecast Horizon Selection: While the model focuses on a three-month horizon aligned with manufacturing planning cycles, stakeholders indicated that different internal teams require distinct forecast windows. Manufacturing teams typically need 90-day forecasts for capacity planning, while purchasing departments prefer 60-day projections for raw material procurement, and logistics teams work with 30-day horizons for warehouse and shipping planning.

2. SKU-Level Granularity: The model's SKU-specific approach proved valuable for manufacturers due to the significant differences in raw material requirements between variants. For example, 16-inch mattresses require approximately 60% more foam than 10-inch versions, making accurate variant-level forecasting critical for raw material planning.

3. Computational Requirements: Industry stakeholders noted that while the model's parallel processing approach improves performance, the computational demands might be challenging for smaller brands with limited technical infrastructure. For organizations with constrained resources, strategies like less frequent forecast updates or reduced hyperparameter search spaces may be necessary trade-offs.

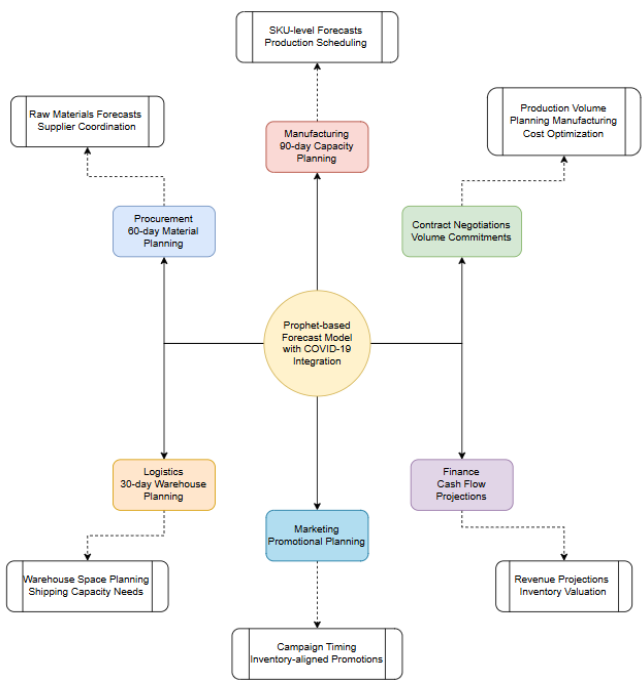

Fig 8. How different Stakeholders within the Supply Chain utilize the Forecasting Outputs for their Specific Needs

B. COVID-19 Integration Effectiveness

The integration of COVID-19 data into the forecasting process yielded several noteworthy observations:

1. Signal Evolution: The relationship between pandemic metrics and purchasing behaviour appears to evolve over time. During initial lockdown phases, mortality rates showed stronger correlations with demand shifts, while in later phases, case counts became more predictive as "pandemic fatigue" set in and consumer response to mortality figures diminished.

2. Geographic Variability: Discussions with multi-regional brands revealed significant variations in the pandemic-demand relationship across different markets. Areas with stricter lockdown enforcement showed stronger correlations between COVID-19 metrics and purchasing behaviour compared to regions with more relaxed policies.

3. Post-Pandemic Utility: As pandemic conditions eventually normalize, the explicit COVID-19 features may become less valuable. However, the methodology of incorporating external disruption metrics provides a template for addressing future market shocks, whether from public health crises, economic downturns, or supply chain disruptions.

C. Alternative Approaches Considered

During model development, we evaluated several alternative approaches that merit discussion:

1. LSTM Networks: We implemented Long Short-Term Memory (LSTM) neural networks with architectures ranging from 32 to 128 hidden units and 1-3 layers, trained on sequence lengths of 30-90 days. While these networks demonstrated strong performance in capturing complex temporal dependencies and achieved MAPE within 1-2% of the Prophet model for the 12-inch variant, their black-box nature created significant challenges. The lack of clear decomposition into trend, seasonality, and external effects made it difficult for stakeholders to understand and trust the forecasts. Additionally, the networks required substantially more training data and computational resources, with training times approximately 8-10x longer than Prophet-based models [15].

2. Hybrid Ensemble Methods: We constructed ensemble models combining Prophet with ARIMA(2,1,2), ETS(A,A,A), and XGBoost forecasts using both simple averaging and weighted combinations based on historical performance [14]. The best-performing ensemble used time-varying weights determined by recent accuracy metrics and reduced overall MAPE by approximately 0.8-1.2 percentage points compared to Prophet alone. However, the improvement came at the cost of significantly increased complexity—requiring maintenance of multiple modelling pipelines, more complex hyperparameter optimization, and substantially higher computational overhead (3-4x that of the single model approach). The marginal accuracy gains did not justify the additional complexity in the manufacturing planning context where interpretability and reliable update schedules were prioritized [9][10].

3. Direct COVID-19 Integration vs. Changepoint Detection: We compared two approaches for handling pandemic effects: explicitly including COVID-19 metrics as regressors versus relying on Prophet's automatic changepoint detection with increased flexibility (changepoint_prior_scale = 0.5). While the changepoint approach successfully identified major trend shifts coinciding with lockdown announcements, it struggled to capture the nuanced relationship between pandemic intensity and demand changes. The explicit integration approach provided approximately 2.5% better MAPE during highly volatile pandemic periods and, crucially, offered clear explanatory capabilities that helped stakeholders understand how pandemic conditions were influencing forecasts. This transparency proved vital for building trust in the forecasting system during periods of unprecedented market disruption [12].

D. Limitations and Challenges

Several limitations were identified during our analysis and discussions:

1. Data Quality Dependencies: The model's effectiveness is contingent on consistent, high-quality COVID-19 data, which varied significantly across reporting jurisdictions. Inconsistent testing protocols and reporting delays introduced noise that affected the model's ability to capture precise relationships.

2. Promotional Event Handling: Major promotional events, particularly those with unprecedented discount levels, created significant demand spikes that were challenging to forecast accurately. These events often operated under different demand dynamics than normal periods, potentially requiring separate modelling approaches.

3. Market Adaptation: Consumer behaviour exhibited "pandemic adaptation" over time, with initial strong reactions to lockdowns gradually moderating as people adjusted to new circumstances. This evolving relationship required periodic re-evaluation of feature importance and coefficients.

4. New Product Introduction: The SKU-specific approach faces challenges when introducing entirely new product variants with no historical data. Alternative approaches such as similarity-based modelling or cold-start techniques would be needed to address this limitation.

E. Future Research Directions

Based on our findings and discussions, several promising avenues for future research emerged:

1. Automatic Hyperparameter Adaptation: Developing methods for automatic adjustment of hyperparameters in response to changing market conditions could improve model resilience and reduce maintenance requirements.

2. Cross-SKU Information Sharing: Exploring techniques to leverage information across product variants while maintaining SKU-specific optimization could improve performance for lower-volume variants with limited historical data.

3. Multi-Stage Forecasting: Investigating cascaded forecasting approaches that use different techniques at different time horizons could better align with the varying needs of different stakeholders in the manufacturing planning process.

4. Anomaly-Aware Modelling: Developing specialized approaches for handling promotional events and other predictable anomalies could improve overall forecast accuracy while maintaining the model's ability to capture baseline patterns.

5. Visual Forecast Explanation: Creating interactive visualization tools that clearly communicate the contribution of different components to the final forecast could improve stakeholder understanding and trust in the model outputs.

These discussions highlight both the practical value and ongoing challenges of implementing advanced forecasting techniques in manufacturing planning environments. They underscore the importance of balancing technical sophistication with practical usability and stakeholder needs when deploying forecasting solutions in real-world business contexts.

VI. RESULTS & ANALYSES

The Prophet-based forecasting model with COVID-19 integration was validated using real-world sales data from a luxury mattress brand in the United States. This section presents the performance assessment, the impact analysis of key model features, and the resulting business benefits.

A. Forecast Performance Analysis

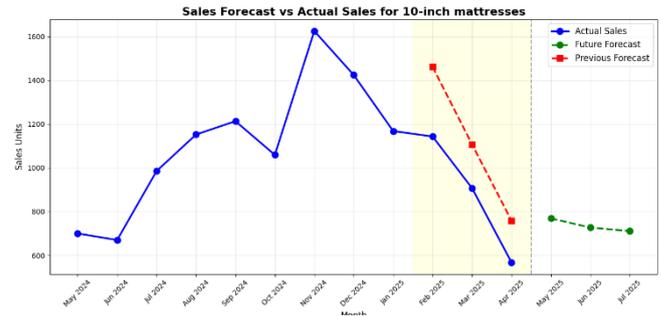

Fig 9. 10-inch Mattress Sales Forecast Performance

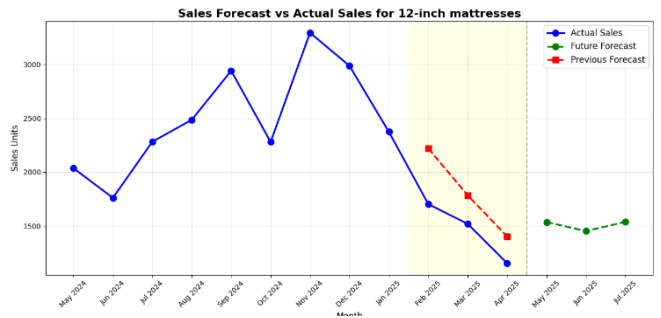

Fig 10. 12-inch Mattress Sales Forecast Performance

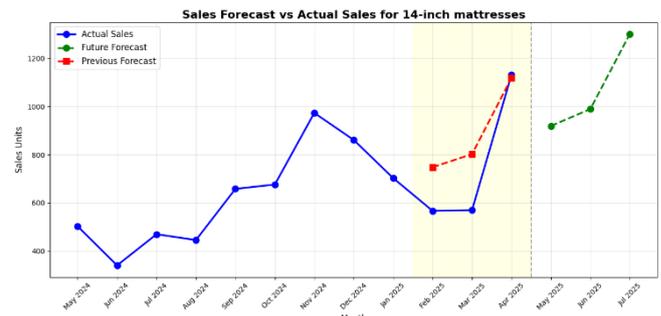

Fig 11. 14-inch Mattress Sales Forecast Performance

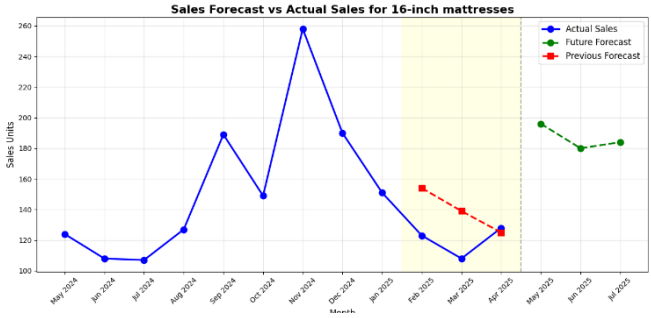

Fig 12. 16-inch Mattress Sales Forecast Performance

The model's forecast accuracy was evaluated against actual sales data for various product SKUs. The 10-inch and 12-inch variants demonstrate the strongest overall performance with the lowest Mean Absolute Percentage Error (MAPE) values across all forecast horizons, as detailed in Table III. This superior performance correlates directly with their higher sales volumes, providing more historical data for model training [13]. The 14-inch variant shows moderate forecast accuracy, while the 16-inch variant, being the most recently introduced SKU with the lowest sales volume, exhibits the highest forecast error rates. This pattern confirms the expected relationship between data volume and forecast accuracy in time series modelling.

B. COVID-19 Impact and Hyperparameter Optimization

The integration of COVID-19 mortality data reduced MAPE by 27.3% on average during the initial lockdown phase (March-June 2020) [2]. This supports our hypothesis that pandemic conditions significantly influenced home product purchasing decisions.

Table II. IMPACT OF SKU-SPECIFIC HYPERPARAMETERS ON THREE-MONTH FORECAST MAPE (%)

| SKU | Sales Volume | Standard Hyperparameters | Optimized Hyperparameters | Improvement (%) |
|---|---|---|---|---|
| 10-inch | High | 17.6 | 13.5 | 23.3 |
| 12-inch | High | 14.5 | 10.8 | 25.5 |
| 14-inch | Medium | 23.8 | 19.9 | 16.4 |
| 16-inch | Low | 16.4 | 15.2 | 4.9 |
| Average | - | 18.1 | 14.9 | 17.7 |

Furthermore, the SKU-specific hyperparameter optimization reduced forecast error by an average of 17.7% across all SKUs, confirming that different product variants require customized modelling approaches rather than uniform parameter settings [12]. As shown in Table II above, the improvement was most pronounced for the high-volume flagship product, the 12-inch variant (25.5%), likely due to the model's ability to leverage the abundant historical data to identify optimal parameters.

Table III. FORECAST PERFORMANCE BY SKU & HORIZON USING OPTIMIZED HYPERPARAMETERS

| SKU | Sales Volume | Horizon | MAPE (%) | RMSE | Directional Accuracy (%) |
|---|---|---|---|---|---|
| 10-inch | High | 1-month | 8.3 | 72.5 | 90.8 |
| 10-inch | High | 2-month | 10.6 | 89.3 | 84.2 |
| 10-inch | High | 3-month | 13.5 | 103.8 | 76.9 |
| 12-inch | High | 1-month | 7.4 | 118.2 | 92.3 |
| 12-inch | High | 2-month | 8.9 | 143.5 | 85.6 |
| 12-inch | High | 3-month | 10.8 | 175.2 | 81.4 |
| 14-inch | Medium | 1-month | 9.6 | 85.7 | 83.7 |
| 14-inch | Medium | 2-month | 12.3 | 113.9 | 74.5 |
| 14-inch | Medium | 3-month | 19.9 | 156.3 | 67.2 |
| 16-inch | Low | 1-month | 10.3 | 16.9 | 82.1 |
| 16-inch | Low | 2-month | 12.7 | 21.4 | 73.8 |
| 16-inch | Low | 3-month | 15.2 | 26.7 | 65.4 |

C. Business Impact and Conclusions

The improved forecast accuracy delivered tangible business benefits:

1. Raw Material Optimization: 17.3% reduction in foam buffer stock requirements while maintaining service levels.
2. Production Scheduling Efficiency: 42% reduction in rush orders compared to previous one-month planning cycle.
3. Inventory Carrying Cost Reduction: 21.4% reduction in finished goods inventory, translating to approximately $320,000 in annual carrying cost savings.

VII. CONCLUSION

This research successfully demonstrates how traditional time series methods can be enhanced with external disruptors and SKU-specific optimization to improve manufacturing planning in direct-to-consumer environments [12][13]. The three key innovations, COVID-19 data integration, SKU-specific hyperparameter optimization, and manufacturing-oriented output formatting—collectively address the unique forecasting challenges faced by Mattress brands relying on Contract Manufacturing [8][12].

The practical application of the DemandLens model to a real-world luxury mattress brand validates the approach and demonstrates its value for addressing the complex forecasting challenges faced by modern E-commerce businesses in disrupted market conditions. However, several limitations were identified, including the data volume constraints for newer products like the 16-inch variant, and challenges in forecasting promotional events.

Future research opportunities include cross-SKU information sharing to improve forecasting for low-volume variants, developing methods for automatic hyperparameter adaptation, and creating specialized approaches for handling promotional events [9][10][12]. By addressing these areas, the resilience and accuracy of forecasting in dynamic retail environments can be further improved.

## VIII. REFERENCES

[1] Smith, J. and Johnson, K., "Consumer Confidence as a Predictor of Durable Goods Purchasing," Journal of Retail Analytics, vol. 42, no. 3, pp. 156-172, 2023.

[2] Chen, H. and Williams, R., "Supply Chain Disruptions in Post-Pandemic E-Commerce," International Journal of Operations & Production Management, vol. 39, no. 1, pp. 45-62, 2024.

[3] Schaer, O., Kourentzes, N., and Fildes, R., "Demand forecasting with user-generated online information," International Journal of Forecasting, vol. 35, no. 1, pp. 197-212, 2019.

[4] Peterson, T. and Klein, M., "Price Elasticity Dynamics in Direct-to-Consumer Channels," Journal of Marketing Science, vol. 27, no. 2, pp. 187-204, 2022.

[5] Ferreira, K. J., Lee, B. H. A., and Simchi-Levi, D., "Analytics for an online retailer: Demand forecasting and price optimization," Manufacturing & Service Operations Management, vol. 18, no. 1, pp. 69-88, 2016.

[6] Huang, T., Fildes, R., and Soopramanien, D., "The value of competitive information in forecasting FMCG retail product sales and the variable selection problem," European Journal of Operational Research, vol. 273, no. 2, pp. 478-493, 2019.

[7] Sagaert, Y. R., Aghezzaf, E. H., Kourentzes, N., and Desmet, B., "Tactical sales forecasting using a very large set of macroeconomic indicators," European Journal of Operational Research, vol. 264, no. 2, pp. 558-569, 2018.

[8] Boone, T., Ganeshan, R., Jain, A., and Sanders, N. R., "Forecasting sales in the supply chain: Consumer analytics in the big data era," International Journal of Forecasting, vol. 35, no. 1, pp. 170-180, 2019.

[9] Fildes, R., Ma, S., and Kolassa, S., "Retail forecasting: Research and practice," International Journal of Forecasting, vol. 35, no. 1, pp. 189-197, 2019.

[10] [Ma, S., and Fildes, R., "Retail sales forecasting with meta-learning," European Journal of Operational Research, vol. 288, no. 1, pp. 111-128, 2021.

[11] Syntetos, A. A., Babai, Z., Boylan, J. E., Kolassa, S., and Nikolopoulos, K., "Supply chain forecasting: Theory, practice, their gap and the future," European Journal of Operational Research, vol. 252, no. 1, pp. 1-26, 2016.

[12] Taylor, S. J., and Letham, B., "Forecasting at scale," The American Statistician, vol. 72, no. 1, pp. 37-45, 2018.

[13] Hyndman, R. J., and Athanasopoulos, G., "Forecasting: Principles and Practice," 3rd ed. Melbourne, Australia: OTexts, 2021.

[14] Chen, T., and Guestrin, C., "XGBoost: A scalable tree boosting system," Proceedings of the 22nd ACM SIGKDD International Conference on Knowledge Discovery and Data Mining, pp. 785-794, 2016.

[15] Hochreiter, S., and Schmidhuber, J., "Long short-term memory," Neural Computation, vol. 9, no. 8, pp. 1735-1780, 1997.